\documentclass[letterpaper, 10 pt, conference]{ieeeconf}  

\IEEEoverridecommandlockouts                              

\usepackage{graphicx} 
\usepackage{amsmath} 
\usepackage{amssymb}  
\usepackage{color,soul}
\usepackage{booktabs}
\usepackage{comment}
\usepackage[normalem]{ulem}
\usepackage[linesnumbered,ruled,vlined]{algorithm2e}
\SetKwInput{KwInput}{Input}               
\SetKwInput{KwOutput}{Output}  

\title{\LARGE \bf LEARNEST: LEARNing Enhanced Model-based State ESTimation for Robots using Knowledge-based Neural Ordinary Differential Equations
}

\author{Kong Yao Chee and M. Ani Hsieh
\thanks{
This work was supported by the Office of Naval Research (ONR) Award No. N00014-22-1-2157 and DSO National Laboratories, 12 Science Park Drive, Singapore 118225.}
\thanks{The authors are with the GRASP Laboratory, University of Pennsylvania, Philadelphia, PA 19104, USA.
        {\tt\small \{ckongyao,\, m.hsieh\}@seas.upenn.edu}}
}

\begin{document}

\maketitle
\thispagestyle{empty}
\pagestyle{empty}

\begin{abstract}
State estimation is an important aspect in many robotics applications. In this work, we consider the task of obtaining accurate state estimates for robotic systems by enhancing the dynamics model used in state estimation algorithms. Existing frameworks such as moving horizon estimation (MHE) and the unscented Kalman filter (UKF) provide the flexibility to incorporate nonlinear dynamics and measurement models. However, this implies that the dynamics model within these algorithms has to be sufficiently accurate in order to warrant the accuracy of the state estimates. To enhance the dynamics models and improve the estimation accuracy, we utilize a deep learning framework known as knowledge-based neural ordinary differential equations (KNODEs). The KNODE framework embeds prior knowledge into the training procedure and synthesizes an accurate hybrid model by fusing a prior first-principles model with a neural ordinary differential equation (NODE) model. In our proposed LEARNEST framework, we integrate the data-driven model into two novel model-based state estimation algorithms, which are denoted as KNODE-MHE and KNODE-UKF. These two algorithms are compared against their conventional counterparts across a number of robotic applications; state estimation for a cartpole system using partial measurements, localization for a ground robot, as well as state estimation for a quadrotor. Through simulations and tests using real-world experimental data, we demonstrate the versatility and efficacy of the proposed learning-enhanced state estimation framework.
\end{abstract}
\section{INTRODUCTION}
\begin{figure} 
    \centering
    {\vspace*{0.3cm}\includegraphics[scale=0.26, trim = 0cm 1cm 0cm 0cm]{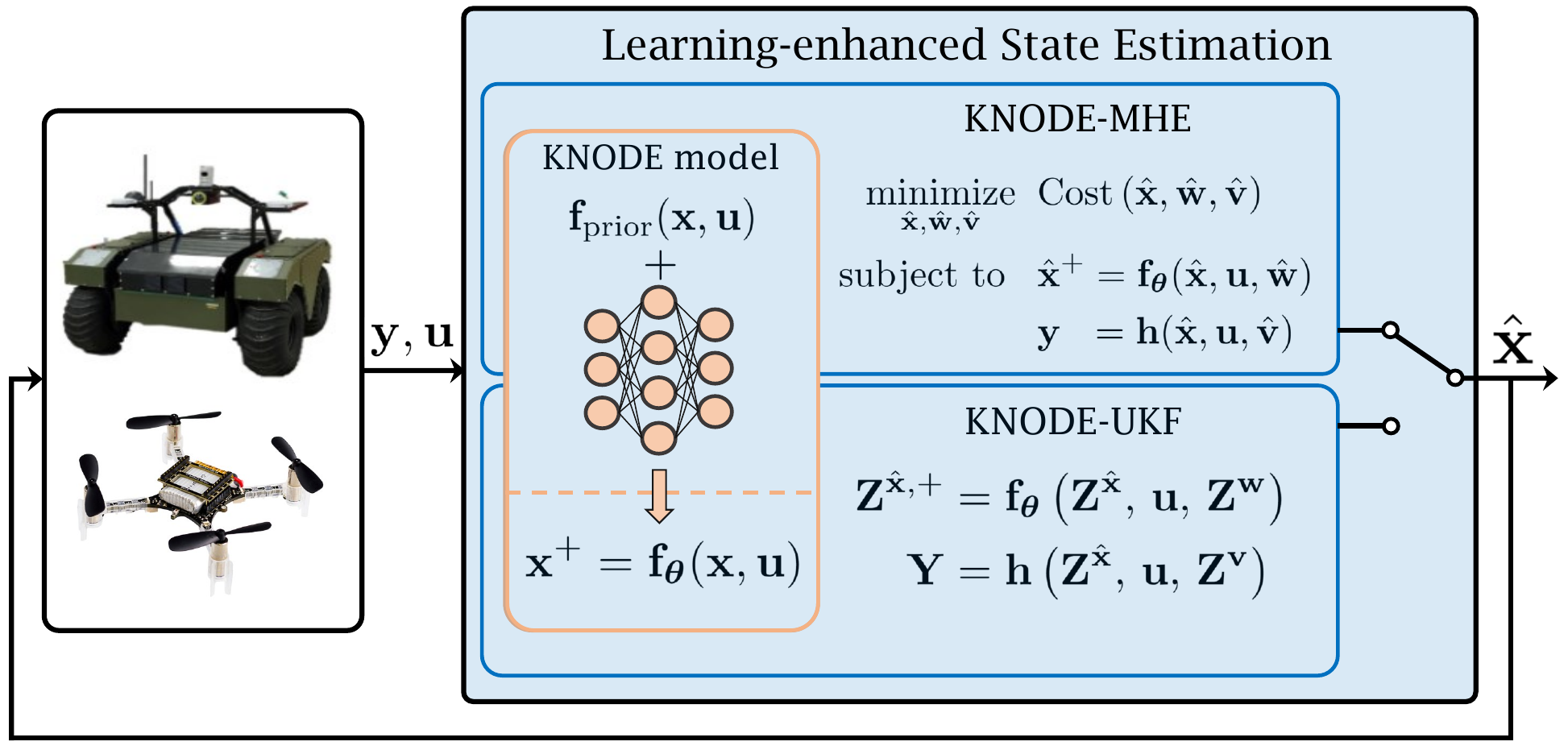}}
    \caption{Schematic of our proposed framework, LEARNEST, applied to a ground robot and a quadrotor system. A prior first-principles model is combined with a neural network to form a hybrid KNODE model. This model is integrated into KNODE-MHE or KNODE-UKF to derive accurate state estimates $\hat{\mathbf{x}}$, based on measurements $\mathbf{y}$ and control inputs $\mathbf{u}$. The feedback path indicates that the state estimates are utilized in other tasks such as motion control or path planning. \textit{Image sources for robots:} \cite{UGV}, \cite{Bitcraze}.}
    \label{fig:overall_framework}
\end{figure}

In many practical applications, it is paramount for robots to acquire accurate information about their translational and rotational dynamics. This information, collectively known as the state, can be obtained through a wide variety of state estimation algorithms \cite{barfoot2017state}. The choice of algorithms often depends on the dynamics of the robot and the sensors used to collect measurements. In applications where the robot dynamics and sensor measurements can be modelled in a linear or quasi-linear manner, algorithms such as the Kalman filter and Extended Kalman filter can be applied. On the other hand, estimation algorithms such as the moving horizon estimator (MHE) and unscented Kalman filter (UKF) allow the integration of nonlinear dynamics and measurement models. Advancements in optimization algorithms and increases in hardware computational power have led to a surge in the use of deep learning tools for robotics applications \cite{pierson2017deep}. These learning tools provide a framework in which data can be leveraged to construct models. With these models, robots are able to obtain more or better information about their states and the environment they are operating in. In this work, we use a deep learning tool, knowledge-based neural ordinary differential equations (KNODEs), to derive accurate dynamics models from data. The KNODE model consists of two components: a prior model derived from first principles and a neural ordinary differential equation that accounts for residual dynamics. In our proposed learning-enhanced state estimation framework, LEARNEST, we apply these accurate knowledge-based, data-driven models into two state estimation algorithms, which in turn improve the accuracy of the state estimates. A schematic of LEARNEST is depicted in Fig. \ref{fig:overall_framework}.

\setlength{\textfloatsep}{8pt}

\section{RELATED WORK}
There are a number of works in the literature that use deep learning tools to learn the dynamics of robotic systems. In \cite{bauersfeld2021neurobem}, the authors use a neural network to learn the residual aerodynamic forces of a quadrotor. The authors in \cite{chee2022knode} model the residual dynamics of an quadrotor system with a KNODE model and use the model for predictive control. O’Connell {\it et al.} \cite{o2022neural} propose an algorithm to learn basis functions that represent the aerodynamics of a quadrotor and use adaptive control to mitigate wind disturbances for a quadcopter. Gaussian processes (GP) regression is another popular tool used to model residual dynamics in robotic applications \cite{torrente2021data, kabzan2019learning}. However, it is well known that GPs suffer from the curse of dimensionality and are unable to fully utilize large amounts of data for both training and inference. In many of these above-mentioned applications, the robots have access to full state measurements and the state estimation problem is not addressed. 

In terms of state estimation, there are a few papers in the literature that leverage the power of deep learning to enhance state estimation algorithms. In \cite{wang2021}, the authors use a neural network to optimize the cost matrices of the MHE optimization problem. This approach, however, does not address the issue of an inaccurate dynamics model. Muntwiler {\it et. al} \cite{muntwiler2022learning} formulate the MHE problem using a convex optimization layer with trainable parameters in the dynamics and measurement models. However, this is only applicable for linear systems and has not been shown to be applicable in physical experiments. The authors in \cite{turner2012} use GPs to learn the parameters in the unscented transformation within the UKF. In \cite{li2020}, the authors use a variant of recurrent neural networks (RNN) to learn the full dynamics of an electric vehicle. The RNN is applied to a UKF framework to predict the vehicle states. In a similar vein, the authors in \cite{kim2021_ukf} propose the combination of a set of LSTM networks into the UKF to filter sensor measurements and estimate the sideslip angle, which is part of the state of a vehicle. 

In contrast to these works, our framework makes use of a combination of first-principles models and neural ODEs to learn {\it hybrid} dynamics models. This integration of first-principles models allows training to be more sample-efficient as the neural network is only required to learn the residual dynamics. Different from RNNs, KNODEs, being lightweight neural networks, do not suffer from exploding and diminishing gradients during training \cite{hochreiter2001gradient, pascanu2013difficulty}. To the best of the authors' knowledge, this is the first work that proposes the integration of a learning-enhanced hybrid model, that combines first principles knowledge with a neural ODE, into both the MHE and UKF algorithms.

Our contributions in this work are three-fold. First, we propose a procedure to train a hybrid knowledge-based data-driven model, known as the KNODE model, using partial or indirect measurements that are possibly nonlinear. This is practical because in many applications, robots do not have access to the full state. Second, we propose a general framework, LEARNEST, in which we integrate the KNODE model into two novel learning-enhanced state estimation algorithms, which we denote as KNODE-MHE and KNODE-UKF. These algorithms are more accurate than their non-learning counterparts in terms of estimation accuracy, as data is assimilated into the estimation process in a systematic and amenable manner, through the synthesis of a KNODE model. Third, we demonstrate the versatility of our framework by considering a range of robotic applications. We provide analysis of the LEARNEST framework through simulations, and verify its capability by applying it onto data collected in physical experiments.

\section{PROBLEM FORMULATION} \label{Problem_section}
We consider robot systems with dynamics that are described in the following form,
\begin{equation} \label{eq:true_dyn}
\begin{split}
    \mathbf{x}^+ = \mathbf{f}(\mathbf{x},\,\mathbf{u},\,\mathbf{w}),\quad \mathbf{y}=\mathbf{h}(\mathbf{x},\,\mathbf{u},\,\mathbf{v}),
\end{split}
\end{equation}
where $\mathbf{x} \in \mathbb{R}^n$ is the state vector, $\mathbf{u} \in \mathbb{R}^p$ is the control input, $\mathbf{y} \in \mathbb{R}^m$ is the measurement vector, and $\mathbf{x}^+ \in \mathbb{R}^n$ is the state at the next time step. The function $\mathbf{f}:\mathbb{R}^n \times \mathbb{R}^p \times \mathbb{R}^q\rightarrow \mathbb{R}^n$ describes the true robot dynamics and $\mathbf{h}:\mathbb{R}^n \times \mathbb{R}^p \times \mathbb{R}^r \rightarrow \mathbb{R}^m$ is a function that maps the state and control inputs to the measurements. The system dynamics and measurements are subjected to process and measurement noises, which are denoted by $\mathbf{w} \in \mathbb{R}^q$ and $\mathbf{v} \in \mathbb{R}^r$. In this setup, the objective is to obtain accurate estimates of the state vector $\mathbf{x}$, given sequences of the measurement vector $\mathbf{y}$ and the control input $\mathbf{u}$.

In a standard state estimation algorithm, a first-principles model $\mathbf{f}_{\text{prior}}$ is used to propagate the dynamics of the system. However, due to uncertainty and model errors, it is unlikely that this model matches the true dynamics perfectly. This results in a degradation of the estimation accuracy. In this work, to improve the accuracy of the dynamics model and state estimates, we propose LEARNEST in which we apply accurate data-driven dynamics models into two learning-enhanced state estimation algorithms, KNODE-MHE and KNODE-UKF. These algorithms provide accurate state estimates by minimizing the discrepancy between the dynamics model and the true system. 

\section{METHODOLOGY}
\subsection{Moving Horizon Estimation} \label{section:MHE}
The MHE framework provides an optimization-based solution to the state estimation problem. At each time step $k$, given the past measurements $\{\mathbf{y}_i\}^k_{i=k-N}$ and control inputs $\{\mathbf{u}_i\}^{k-1}_{i=k-N}$, the following nonlinear optimization problem is formulated and solved over a moving horizon of length $N$ \cite{rao2003constrained},
\begin{equation} \label{eq:mhe}
\begin{split}
    \underset{\{\hat{\mathbf{x}}_{i|k}\},\{\hat{\mathbf{w}}_{i|k}\},\{\hat{\mathbf{v}}_{i|k}\}}{\textnormal{minimize}}\;\;\; & ||\hat{\mathbf{x}}_{k-N|k} -\Bar{\mathbf{x}}_{k-N|k} ||^2_{\mathbf{P}_k} + \\
    &\sum_{i=k-N}^{k-1} ||\hat{\mathbf{w}}_{i|k}||^2_\mathbf{Q} +\sum_{i=k-N}^{k} ||\hat{\mathbf{v}}_{i|k}||^2_\mathbf{R}\\
    \text{subject to}\;\;\; \hat{\mathbf{x}}_{i+1|k} &= \mathbf{f}(\hat{\mathbf{x}}_{i|k}, \mathbf{u}_{i}, \hat{\mathbf{w}}_{i|k}), \\&\qquad\qquad \forall\, i=k-N,\dotsc,k-1,\\
    \; \mathbf{y}_{i} &= \mathbf{h}(\hat{\mathbf{x}}_{i|k}, \mathbf{u}_{i}, \hat{\mathbf{v}}_{i|k}), 
    \\&\qquad\qquad \forall\, i=k-N,\dotsc,k,
\end{split}
\end{equation}
where $\{\hat{\mathbf{x}}_{i|k}\}^k_{i=k-N},\{\hat{\mathbf{w}}_{i|k}\}^{k-1}_{i=k-N}$ and $\{\hat{\mathbf{v}}_{i|k}\}^k_{i=k-N}$ denote the sequences of estimated states, process, and measurement noises respectively. In addition, $||\mathbf{s}||^2_A$ denotes $\mathbf{s}^T A \mathbf{s}$, $\{\mathbf{s}_i\}^u_{i=l}$ represents the sequence $\{\mathbf{s}_l,\dotsc,\mathbf{s}_u\}$, and $\mathbf{Q}$ and $\mathbf{R}$ are the cost matrices that penalize the influence of the process and measurement noises on the objective function. The cost matrix $\mathbf{P}_k$ accounts for past state information. The term $\Bar{\mathbf{x}}_{k-N|k}$ is the \textit{a priori} estimate for the initial state $\hat{\mathbf{x}}_{k-N|k}$. One way to update $\Bar{x}_{k-N|k}$ is to use the optimal state estimate obtained from the previous time step, \textit{i.e.,} $\Bar{\mathbf{x}}_{k-N|k} \leftarrow \hat{\mathbf{x}}^{\star}_{k-N|k-1}$ \cite{alessandri2008moving}. More details on this formulation can be found in \cite{rao2003constrained, robertson1996moving}.  

At each time step $k$, upon solving the optimization problem \eqref{eq:mhe}, the last element of the sequence of optimal estimated states, $\hat{\mathbf{x}}^{\star}_{k|k} \in \{\hat{\mathbf{x}}^{\star}_{i|k}\}_{i=k-N}^k$ is applied as the current state estimate. Notice that the constraints in \eqref{eq:mhe} consist of a dynamics model $\mathbf{f}$. In a standard MHE formulation, a prior model $\mathbf{f}_{\text{prior}}(\mathbf{x},\mathbf{u})$ is derived from first-principles knowledge of the system and used as the dynamics model within \eqref{eq:mhe}. In the case where the model is accurate, this procedure naturally gives accurate state estimates. However, it is likely that there are residual dynamics or uncertainty that are not accounted for in these models. Therefore, in this work, we apply the KNODE framework to account for these uncertainties. Instead of only using a prior model, we use a {\it hybrid} dynamics model, denoted by $\mathbf{f}_{h}(\mathbf{x},\mathbf{u})$, obtained through the KNODE framework (see Sections \ref{section:KNODE} and \ref{section:KNODE_training}). The objective is to learn $\mathbf{f}_{h}(\mathbf{x},\mathbf{u})$ such that it is a more accurate representation of the true system dynamics \eqref{eq:true_dyn}. This in turn allows us to integrate it into the KNODE-MHE framework in LEARNEST to provide more accurate state estimates than a conventional MHE formulation.

\subsection{Unscented Kalman Filter} \label{section:UKF}
The UKF is a variant of the Kalman filter \cite{julier1997new, wan2000unscented}. As an extension of the extended Kalman filter, it allows the direct integration of nonlinear dynamics models without linearization. This ability to integrate nonlinear dynamics models enables us to assimilate the hybrid model learned from the KNODE framework into the UKF. At each time step $k$, the UKF algorithm consists of three main steps; computing the sigma vectors based on the previous estimate, performing a time update of the sigma vectors using the dynamics and measurement models, and lastly, updating the state estimate using the collected measurements \cite{wan2000unscented}. By defining an augmented vector $\mathbf{z} :=[\mathbf{x}^{\top}\; \mathbf{w}^{\top}\; \mathbf{v}^{\top}]^{\top}$ and by combining the sigma vectors computed through the unscented transformation into a matrix $\mathbf{Z} := [\mathbf{Z}^{\mathbf{x}\top}\; \mathbf{Z}^{\mathbf{w}\top}\; \mathbf{Z}^{\mathbf{v}\top}]^{\top} := \left[\hat{\mathbf{z}} \;\,\hat{\mathbf{z}} \pm \sqrt{(l+\lambda)\mathbf{P}^{\mathbf{z}}}\right]$, the equations for the time update are given as \cite{wan2000unscented},
\begin{equation} \label{eq:ukf_time}
\begin{split}
    \mathbf{Z}^{\mathbf{x}}_{+} &= \mathbf{f}\left(\mathbf{Z}^{\mathbf{x}},\, \mathbf{u},\, \mathbf{Z}^{\mathbf{w}}\right),\quad
    \hat{\mathbf{z}}^{-} = \sum_{i=0}^{2l} {W}_{m,i} \mathbf{Z}^{\mathbf{x}}_{+,i}, \\
    \mathbf{Y} &= \mathbf{h}\left(\mathbf{Z}^{\mathbf{x}},\, \mathbf{u},\, \mathbf{Z}^{\mathbf{v}}\right),\quad
    \hat{\mathbf{y}}^{-} = \sum_{i=0}^{2l} {W}_{m,i} \mathbf{Y}_{i}, \\
    \mathbf{P}^{-} &= \sum_{i=0}^{2l} {W}_{c,i} \left(\mathbf{Z}^{\mathbf{x}}_{i,+} - \hat{\mathbf{z}}^{-}\right)\left(\mathbf{Z}^{\mathbf{x}}_{i,+} - \hat{\mathbf{z}}^{-}\right)^{\top},\\
\end{split}
\end{equation}
where $l=2n+m$ is the dimension of the augmented vector, $\lambda$ is a scaling parameter and $\mathbf{P}^{\mathbf{z}}$ is the block diagonal covariance matrix of the augmented vector. $\{{W}_{m,i}\}_{i=0}^{2l}$ and $\{{W}_{c,i}\}_{i=0}^{2l}$ are weights computed by the unscented transformation. Details on the computation of the sigma vectors and the measurement update can be found in \cite{wan2000unscented}. 

In \eqref{eq:ukf_time}, a prediction model $\mathbf{f}$ is used to propagate the sigma vectors ahead in time. This model takes on a similar role as the dynamics model within the MHE framework described in Section \ref{section:MHE}. If this dynamics model represents the true system sufficiently well, then the state estimates generated by the UKF will be accurate. In practice, due to uncertainty and unmodelled dynamics, it is unlikely for the model to match the true dynamics exactly. Hence, in the KNODE-UKF algorithm within the LEARNEST framework, we introduce an enhancement, where we replace the prediction model $\mathbf{f}$ with a learned hybrid KNODE model $\mathbf{f}_h$ that represents the true system dynamics with a higher accuracy.

\subsection{Knowledge-based Neural ODEs} \label{section:KNODE}
While the dynamics model derived from prior knowledge of the system achieves a certain level of fidelity, it is generally not sufficient for high-performance model-based applications, where a more accurate model is desired. In robotic systems, uncertainty and residual dynamics manifest in different forms \cite{uncertainModel2, uncertainEnv}. There could be uncertainty in the system properties, or additional perturbations due to aerodynamics, friction, and interactions with the environment. Furthermore, it is often difficult to pinpoint the exact sources of these uncertainties and this makes it challenging to account for their effects in the prior dynamics model. The KNODE framework \cite{jiahao2021knowledgebased} alleviates this issue by learning these uncertainties and residual dynamics through a data-driven procedure. The KNODE framework first considers a continuous-time representation of the true system dynamics,
\begin{equation} \label{eq:cts_model}
\dot{\mathbf{x}} = \mathbf{f}_c(\mathbf{x},\mathbf{u}) := \mathbf{f}_{\text{prior}}(\mathbf{x},\mathbf{u}) + \Delta(\mathbf{x},\mathbf{u}),
\end{equation}
where $\mathbf{f}_c(\mathbf{x},\mathbf{u})$ represents the true dynamics of the system and $\Delta(\mathbf{x},\mathbf{u})$ represents residual dynamics that are not captured by the prior dynamics model $\mathbf{f}_{\text{prior}}(\mathbf{x},\mathbf{u})$. The subscript $c$ is used to denote the continuous-time nature of the model. The residual dynamics $\Delta(\mathbf{x},\mathbf{u})$ is parameterized as a neural ODE with parameters $\boldsymbol{\theta}$, denoted as $\Delta_{\boldsymbol{\theta}}(\mathbf{x},\mathbf{u})$. In other words, the neural ODE is a neural network that approximates the vector field characterizing the residual dynamics. After training, we obtain a hybrid model, 
\begin{equation} \label{eq:cts_hybrid_model}
\mathbf{f}_{h,c}(\mathbf{x},\mathbf{u}) := \mathbf{f}_{\text{prior}}(\mathbf{x},\mathbf{u}) + \Delta_{\boldsymbol{\theta}^{\star}}(\mathbf{x},\mathbf{u}),     
\end{equation}
where we use $^{\star}$ to denote the optimal parameters. We then discretize this model using standard numerical solvers such as the explicit Runge-Kutta $4^{\text{th}}$ order method (RK4) to obtain the hybrid model, $\mathbf{f}_{h}(\mathbf{x},\mathbf{u}) := RK4\left(\mathbf{f}_{h,c}\left(\mathbf{x},\mathbf{u}\right)\right)$. This model is then incorporated into the MHE and UKF frameworks as described in Sections \ref{section:MHE} and \ref{section:UKF}. By incorporating prior models, the learned neural networks are typically lightweight feed-forward networks, which significantly reduce computational time, both during training and inference. More details on the network architectures are given in Sections \ref{section:implementation} and \ref{section:Simulations}.

\subsection{KNODE Training with Partial or Indirect Observations} \label{section:KNODE_training}
The procedure to train a hybrid KNODE model using partial or indirect observations of the state is described in Phase 1 of Algorithm \ref{alg:algo1}. Conceptually, the procedure is similar to the training procedure described in \cite{chee2022knode}, with the exception of Step 1 in Algorithm \ref{alg:algo1}, where the pre-training observer is applied. In \cite{chee2022knode}, measurements of the full state are assumed to be available. In this work, we relax this assumption and consider the case where only partial or indirect measurements are given. To get access to state observations required for training, we propose the addition of a pre-training observer. The pre-training observer $\mathbf{g}$ is a function that maps input and output measurements, sampled at times $\{t_i\}_{i=1}^M$, to state observations and is expressed as
\begin{equation} \label{eq:pre_train_observer}
    \boldsymbol{\zeta}_{1},\dotsc,\boldsymbol{\zeta}_{M} = \mathbf{g}(\mathbf{y}_{1},\dotsc,\mathbf{y}_{M},\mathbf{u}_{1},\dotsc,\mathbf{u}_{M}),
\end{equation}
where $\left\{\boldsymbol{\zeta}_{i}\right\}^M_{i=1}$ are state observations used for training. After applying the pre-training observer, we collect the state observations, together with the control inputs into a dataset $\mathcal{O}$ and compute one-step state predictions,
\begin{equation} \label{eq:one_step}
    \Hat{\boldsymbol{\zeta}}_{{i+1}} = \boldsymbol{\zeta}_{i} +  \int^{t_{i+1}}_{t_i} \mathbf{f}_{\text{prior}}(\boldsymbol{\zeta}_{i},\mathbf{u}_{i}) + \Delta_{\theta}(\boldsymbol{\zeta}_{i}, \mathbf{u}_{i})\, dt.
\end{equation}
With these one-step predictions $\{\Hat{\boldsymbol{\zeta}}_{{i}}\}^M_{i=2}$, we define a mean-squared error loss by considering the deviation between the state observations and one-step predictions,
\begin{equation} \label{eq:loss}
    \mathcal{L}(\theta) := \frac{1}{M-1}\sum^{M}_{i=2}\left\|\hat{\boldsymbol{\zeta}}_{i} - {\boldsymbol{\zeta}}_{i}\right\|^2_2
\end{equation}
From \eqref{eq:one_step} and \eqref{eq:loss}, notice that the prior model $f_{\text{prior}}\left(\boldsymbol{\zeta}_{i},\mathbf{u}_{i}\right)$ is embedded into the loss function through an integral. This embedding allows the KNODE framework to achieve sufficiently high accuracy, without the need to access to the true underlying vector field, which is often noisy or inaccessible, especially in robotics applications. By computing the gradients of the loss function with respect to the parameters $\boldsymbol{\theta}$, we update the parameters in an iterative manner through a standard back-propagation procedure \cite{paszke2017automatic}. With the optimal set of parameters $\boldsymbol{\theta}^{\star}$, we construct the discrete-time hybrid model as described in Section \ref{section:KNODE} and apply it to the proposed KNODE-MHE and KNODE-UKF algorithms.
\setlength{\algomargin}{13pt}
\begin{algorithm} \label{alg:algo1}
\DontPrintSemicolon
  \KwInput{Measurements $\{\mathbf{y}_{i}\}_{i=1}^M$ and control inputs $\{\mathbf{u}_{i}\}_{i=1}^M$ sampled at times $\{t_i\}_{i=1}^M$, prior dynamics model $\mathbf{f}_{\text{prior}}$, pre-training observer $\mathbf{g}$}
  \KwOutput{State estimates $\hat{\mathbf{x}}$, discrete-time KNODE model $\mathbf{f}_h$}
  \textit{Phase 1 (offline):}\;
  \Indp Apply $\mathbf{g}$ in \eqref{eq:pre_train_observer} to get state observations $\{\boldsymbol{\zeta}_{i}\}_{i=1}^M$\;
  Collect dataset $\mathcal{O} := \{{\boldsymbol{\zeta}}_{i}, \mathbf{u}_{i}\}_{i=1}^M$\;
  Compute one-step predictions $\Hat{\boldsymbol{\zeta}}_{{i}}$ using \eqref{eq:one_step}\;
  Compute loss $\mathcal{L}(\theta)$ with \eqref{eq:loss}\;
  Train to get optimal set of parameters $\theta^{\star}$\;
  Construct hybrid model $\mathbf{f}_{h,c}$ with \eqref{eq:cts_hybrid_model}\;
  Discretize to get $\mathbf{f}_h \leftarrow RK4(\mathbf{f}_{h,c})$\;
  Set $\mathbf{f} \leftarrow \mathbf{f}_h$ in \eqref{eq:mhe} or \eqref{eq:ukf_time} for KNODE-MHE or KNODE-UKF\;
  \Indm\textit{Phase 2 (online), at each time step $k$}:\;
  \Indp Collect measurements $\mathbf{y}(k)$ and controls $\mathbf{u}(k)$\;
  Apply KNODE-MHE or KNODE-UKF to get state estimates $\hat{\mathbf{x}}(k)$
\caption{Learning Enhanced Model-based State Estimation Framework, LEARNEST}
\end{algorithm}

\subsection{Implementation Details} \label{section:implementation}
The nonlinear optimization problem within the MHE and KNODE-MHE frameworks is formulated in CasADi \cite{Andersson2019}. An interior-point method solver IPOPT \cite{wachter2006implementation} is used to solve the optimization problem. To improve efficiency, the solver is warm-started at each time step by providing it with an initial guess of the solution, based on the optimal solution obtained from the previous time step. For the implementation of the Kalman filters, we adopt tools in the FilterPy library \cite{FilterPy} and make modifications to accommodate for the hybrid KNODE model. The KNODE training procedure is implemented using the Python-based torchdiffeq library \cite{chen2018neuralode}. The explicit fourth order Runge-Kutta method is used for training. We use Adam \cite{Kingma2015AdamAM} as the optimizer to find the optimal set of parameters $\boldsymbol{\theta}^{\star}$ during the training procedure. The explicit RK4 method is used to discretize $\mathbf{f}_{h,c}$ to get the final hybrid model, $\mathbf{f}_h$, which is integrated into KNODE-MHE and KNODE-UKF. 

\section{SIMULATIONS AND EXPERIMENTS} \label{section:Simulations}
In our experiments and evaluation procedure, we seek to answer these questions pertinent to the LEARNEST framework: (a) How accurate are the learned hybrid models, compared to the models learned with full state information, and against the ground truth, when applied to state estimation algorithms? (b) How much improvement do the KNODE-MHE and KNODE-UKF frameworks provide as compared to the standard MHE and UKF frameworks? (c) Is the framework applicable to real-world data collected from physical experiments of robotic systems? To address the first two questions, we consider two estimation tasks; state estimation for a cartpole given partial state measurements, as well as localization of a ground robot using indirect measurements from both external and onboard sensors. For the third question, we apply the LEARNEST framework on flight data collected from a quadrotor system to estimate its states. 

\subsection{State Estimation for a Cartpole System} \label{section:cartpole}
Consider a cartpole system with the following dynamics \cite{barto1983, florian2007correct},
\begin{equation} \label{eq:cartpole_eqns}
\begin{split}
    \ddot{\alpha} &= \frac{g\sin\alpha-\cos\alpha\left(F+m_p l \dot{\alpha}^2 \sin\alpha \right)}{l\left(\frac{4}{3}-\frac{m_p\cos^2\alpha}{m_c+m_p}\right)},\\
    \ddot{p} &= \frac{F+m_p l \left(\dot{\alpha}^2 \sin\alpha - \ddot{\alpha}\cos\alpha\right)}{m_c+m_p},
\end{split}
\end{equation}

where $p$ is the position of the cart and $\alpha$ is the angle between the pole and the vertical. The cart and the pole have masses $m_c$ and $m_p$ and the pole has a length of $2l$. $g$ is the gravitational force and $F$ is the force acting on the cart, which is the control input acting on the system. A schematic diagram of a cartpole system is shown in Fig. \ref{fig:schematic_systems}. By defining the state vector $\mathbf{x}:=[p \;\dot{p}\; \alpha\; \dot{\alpha}]^{\top}$, these dynamics are discretized and written in the form of \eqref{eq:true_dyn}. To demonstrate the efficacy of the framework, it is assumed that only the position of the cart $p$ and the angle $\alpha$ are measured. In other words, we obtain partial measurements of the state, and the measurement model is given by $\mathbf{y} = C\mathbf{x}$, where $C:=[1\; 0\; 1\; 0]$. In Section \ref{section:robot_localization}, we consider a more challenging, yet practical application, where nonlinear and indirect measurements of the state are provided.

The true cartpole system is numerically simulated using an explicit 5$^{\text{th}}$ order Runge-Kutta method (RK45). In terms of uncertainty, it is assumed that the cart has a true mass of 1.5kg, while the mass of the cart considered in the prior dynamics model $\mathbf{f}_{\text{prior}}$ within the KNODE model is 1kg. Since the mass of the cart $m_c$ is tightly coupled in the equations of motion as shown in \eqref{eq:cartpole_eqns}, this allows us to ascertain the effectiveness of the KNODE framework in terms of modeling residual dynamics. Additionally, we inject zero-mean Gaussian noise with standard deviation of 0.01 across all states, and into the partial state measurements. 

For training, we collect $M=5000$ data points $\{\mathbf{y}_{i},\,\mathbf{u}_{i}\}$ in a single trajectory over a time span of 10 seconds. We implement a KF as the pre-training observer to account for the translational and rotational dynamics and this provides the state observations required for training. The neural ODE architecture consists of 2 layers with 8 hidden neurons in between and uses the hyperbolic tangent as the activation function. The network is trained over 250 epochs and uses the explicit RK4 solver during the training procedure. Training is done on a Intel i5 CPU and was completed in 3 seconds.

\subsection{Localization for a Ground Robot}  \label{section:robot_localization} 
Next, we consider a task of localizing a ground robot using onboard measurements, combined with measurements from multiple external static sensors. This task is more challenging than state estimation for the cartpole system. First, the measurements from the external sensors are nonlinear functions of the state. Furthermore, this application consists of a sensor fusion sub-task, where measurements from multiple sensors are fused together with onboard measurements. Adapted from \cite{rajamani2011vehicle}, we model the dynamics of the ground robot as 
\begin{equation} \label{eq:grd_robot_eqns}
\begin{split}
    \dot{x} &= v\cos(\psi + \beta),\;\;\;\;
    \dot{y} = v\sin(\psi + \beta),\\
    \dot{v} &= a,\quad\qquad\qquad\;\;\;
    \dot{\psi} = \frac{v\cos\beta\tan\delta}{l_f + l_r},
\end{split}
\end{equation}
where $\beta = \tan^{-1}\left((l_r\tan\delta) / (l_f+l_r) \right)$ is the slip angle between the robot velocity vector and its center line. We denote $(x,\,y)$ as the position of the robot in the world frame, $v$ is the speed, and $\psi$ is the heading angle of the robot with respect to the world frame. Additionally, $a$ and $\delta$ are the acceleration and front steering angle of the robot, which act as the control inputs $\mathbf{u}$ to the system. The variables $l_f$ and $l_r$ are the distances from the front and rear wheels to the centre of gravity. The state vector $\mathbf{x}$ is defined as $[x\;y\;v\;\psi]^{\top}$. Each of the external sensors, denoted with index $i$, with position $(p_{i,x},\,p_{i,y})$, provides range measurements $r_i := \left((x-p_{i,x})^2 + (y-p_{i,y})^2\right)^{1/2}$ to the ground robot. The ground robot is assumed to have an onboard sensor that measures its orientation, which in this case is the heading angle $\psi$. This system configuration is typical in robotic systems. In many applications, the robot has onboard sensors that provide partial measurements of its state and simultaneously, has access to some data or information from some external source such as the Global Positioning System (GPS) or other sensors \cite{alatise2020review, chee201366}. 

\begin{figure}
    \centering
    {\vspace*{0.3cm}\includegraphics[scale=0.275, trim = 0cm 1cm 0cm 0cm]{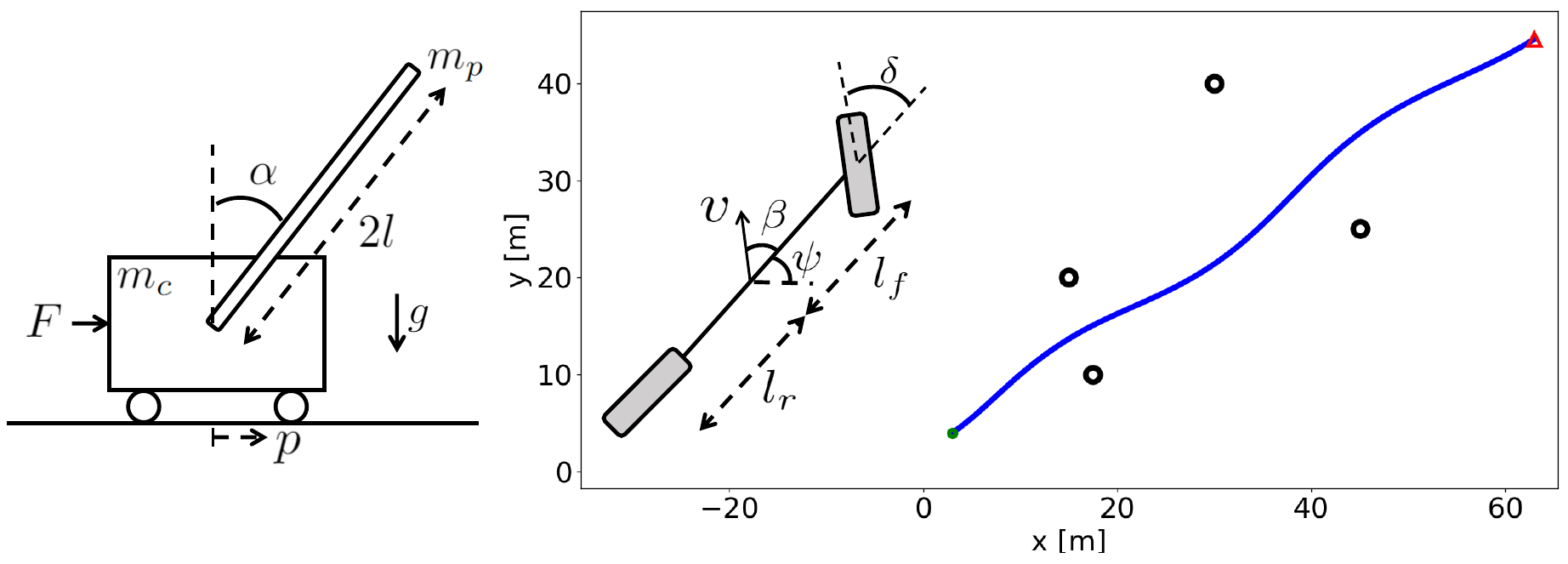}}
    \caption{\textit{Left:} Schematic diagram for the cartpole system. \textit{Right:} Schematic diagram of the ground robot, with its trajectory plotted in blue. The sensors providing measurements to the robot are plotted with black circles. The starting and ending positions of the robot are denoted by a green circle and a red triangle respectively.}
    \label{fig:schematic_systems}
\end{figure}

In our simulations, we consider four sensors providing range measurements to the ground robot. A schematic of the location of the sensors (black circles) and the trajectory of the ground robot (in blue) is shown on the right of Fig. \ref{fig:schematic_systems}. We consider residual dynamics in the form of aerodynamic drag acting on the robot. It is modelled as $a_{\text{drag}} = 0.5SC_Dv^2/m$, where $S$ is the reference area, $C_D$ is the drag coefficient and $m$ is the mass of the robot. In practice, the robot may also experience other forces such as friction or forces depending on the environment it is operating in. In addition to the residual dynamics, zero-mean Gaussian noise with standard deviation of $[0.005\text{m}, 0.005\text{m}, 0.005\text{m/s}, 0.1\text{deg}]$ are added to the states. Gaussian noise with standard deviation of $0.05\text{m}$ and $1.0\text{deg}$ are added to the range and heading measurements respectively. 

For training, we collect $M=2500$ data points in a single trajectory over a time span of 25 seconds. For the pre-training observer, we first apply a precursory UKF to extract the position and heading angle. We then use a linear KF to extract the robot speed from the position observations. The prior model $\mathbf{f}_{\text{prior}}$ used in the KNODE model is obtained from \eqref{eq:grd_robot_eqns}. The neural ODE consists of 2 layers with 8 hidden neurons in between and uses the hyperbolic tangent as the activation function. The network is trained over 10000 epochs and the explicit RK4 solver is used during training.
\setlength{\textfloatsep}{4pt}
\vspace{-0.4cm} 
\subsection{State Estimation for the Crazyflie Quadrotor} \label{section:crazyflie}
In this set of experiments, we use the LEARNEST framework to estimate the translational states of the Crazyflie quadrotor system. Measurements are collected when the open-source Crazyflie 2.1 \cite{Bitcraze} is flown in an indoor environment. A schematic of the experimental setup is depicted on the left of Fig. \ref{fig:crazyflie_results}. The Crazyflie is commanded to follow a circular trajectory with a radius of 0.5m and at a speed of 0.5m/s. The 3-dimensional position measurements are obtained from a VICON motion capture system, while the accelerations are measured from a 3-axis accelerometer onboard the Crazyflie. Given these position and acceleration measurements, the objectives of the state estimation algorithms are to generate accurate estimates of the velocities and to provide noise attenuation for the measurements. The state is defined as $\mathbf{x}:=[\mathbf{p}^{\top} \dot{\mathbf{p}}^{\top} \ddot{\mathbf{p}}^{\top}]^{\top}$, where $\mathbf{p}$ is the position vector of the quadrotor. The prior dynamics model in the KNODE model is represented as $\mathbf{f}_{\text{prior}}(\mathbf{x}) = \mathbf{A}\mathbf{x}$, where $\mathbf{A} \in \mathbb{R}^{9\times9}$ maps the translational states to their derivatives. Even though this is a linear map, the KNODE model also consists of a neural network, which makes the overall dynamics model nonlinear. Ten sets of flight data are collected, with an approximate sampling rate of 100Hz. Each of them has a duration of about 14 seconds. The neural ODE used in this setup consists of 2 layers with 16 neurons in between and uses the hyperbolic tangent as the activation function. A KF is implemented as the pre-training observer and only one out of the 10 datasets is used for training. The performance of the state estimation algorithms on the remaining datasets ascertains the generalization ability of the learned model. Training of the KNODE model is done over 150 epochs and was completed in 2 seconds. 

\section{RESULTS AND DISCUSSION}  \label{section:results_discussion} 
For the tasks described in Sections \ref{section:cartpole} and \ref{section:robot_localization}, we ran 15 simulations for each task and method across different random seeds, with a total of 240 runs. Statistics of the overall mean-squared errors (MSE) between the estimated and true states under KNODE-MHE and KNODE-UKF, against various baselines are summarized in Table \ref{table:sim_results}. The mean and standard deviation of the MSEs are denoted by $\mu$ and $\sigma$ respectively. MHE and UKF are the nominal state estimation algorithms with no learning enhancements. In other words, only the prior model $\mathbf{f}_{\text{prior}}$ is used as the dynamics model in these two frameworks. To understand the effect of the residual dynamics, another set of experiments is conducted in which the dynamics of the true system $\mathbf{f}$ is used as the dynamics model. These results are shown under the rows, MHE (true dynamics) and UKF (true dynamics). These can be interpreted as ideal baselines that are free from any errors induced by the learning procedure, and without any unknown dynamics. The results under KNODE-MHE (full state) and KNODE-UKF (full state) are for the case in which full state measurements are available and used for training. These results, in contrast with those under KNODE-MHE and KNODE-UKF, would ascertain the accuracy of the KNODE model learned under partial or indirect measurements. 

As observed in Table \ref{table:sim_results}, having a more accurate dynamics model significantly improves the accuracy of the state estimates, in both the mean and standard deviation. This can be seen from the comparison between KNODE-MHE, KNODE-UKF and their non-learning counterparts, MHE and UKF. The results imply that learning a dynamics model will improve accuracy of the state estimates, even with partial or indirect measurements. In the case where full state measurements are available, a better model can be obtained and we can expect improvements over the case where partial or indirect measurements are used for learning. It is also observed that the results for KNODE-MHE (full state) and MHE (true dynamics) in the cartpole system are comparable, within $0.3\%$. This implies that in this setting, the accuracy of the KNODE model learned under full state measurements is close to that of the true dynamics model. For both estimation tasks, MHE performs better than UKF, when the residual dynamics are present, but not accounted for.
\vspace{-0.25cm} 
\begin{table}[!htbp]
\centering
\caption{MSE Statistics under 2 State Estimation Tasks. \label{table:sim_results}}
\vspace{-0.25cm} 
\resizebox{\columnwidth}{!}{
\begin{tabular}{*5c}
\toprule
Method &  \multicolumn{2}{c}{Cartpole State Estimation} & \multicolumn{2}{c}{Robot Localization}\\
\midrule
{}   & $\mu$ ($10^{-3}$)   & $\sigma$ ($10^{-3}$)   & $\mu$ ($10^{-3}$)   & $\sigma$ ($10^{-3}$)\\
MHE   &  7.436	& 0.110 & 3.169 & 1.139 \\
KNODE-MHE   &  0.389 & 0.040	&2.046	&1.650 \\
KNODE-MHE (full state)   &  0.339	&0.038	&1.341	&1.133 \\
MHE (true dynamics)   &  0.340	&0.038	&1.306	&1.137\\
\midrule
UKF   &  43.660	&0.187	&69.717	&0.060\\
KNODE-UKF   &  0.0704	&0.0011	&5.059	&0.051\\
KNODE-UKF (full state)   &  0.002	&0.0004	&0.365	&0.0257\\
UKF (true dynamics)   &  0.0007	&0.0002	&0.285	&0.0266\\
\bottomrule
\end{tabular}
}
\end{table}

\begin{figure} 
    \centering
    {\vspace*{0.25cm}\includegraphics[scale=0.265, trim={0cm 1cm 0cm 0cm}]{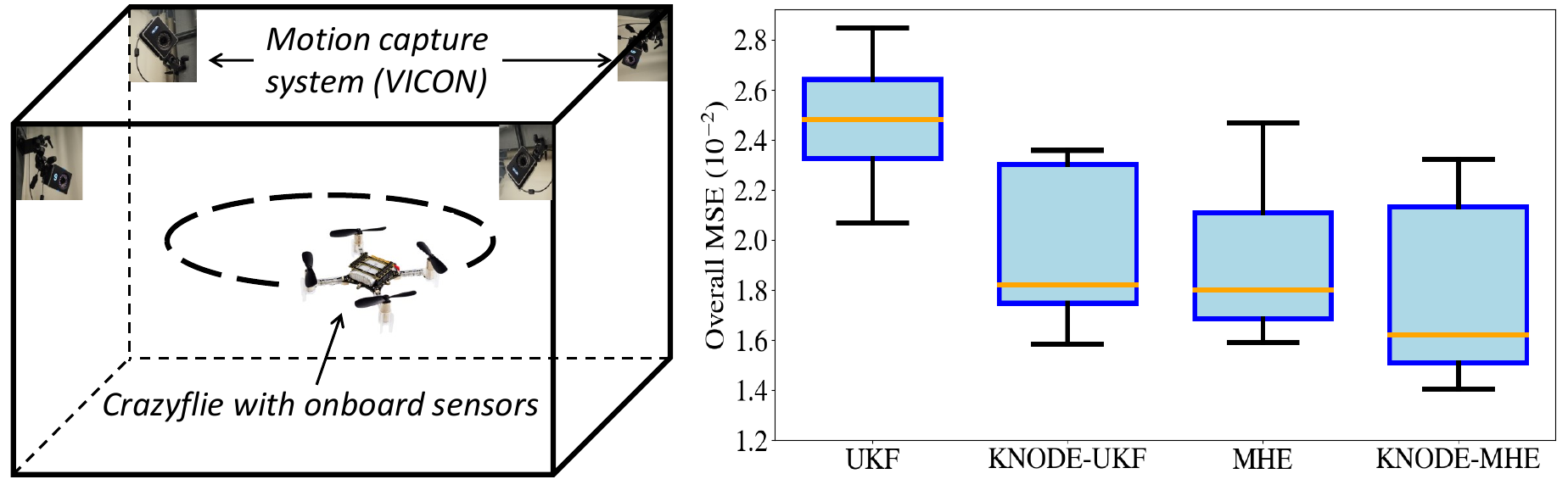}}
    \caption{\textit{Left:} Experimental setup for the Crazyflie system. Partial measurements of the state are obtained from the motion capture system and onboard sensors. \textit{Right:} Statistics of the overall MSE computed over the translational states of the Crazyflie, and across 10 sets of flight data. The orange lines indicate the medians for each method.}
    \label{fig:crazyflie_results}
\end{figure}
\vspace{-0.2cm} 
As described in Section \ref{section:crazyflie}, to ascertain the practicality and efficacy of LEARNEST on real-world data, we apply the framework to 10 sets of flight data collected with the Crazyflie. A Savitzky–Golay filter \cite{schafer2011savitzky} with a window length of 150 and a polynomial of order 2 is applied on the accurate position measurements from VICON to compute velocity and acceleration baselines to compare against the state estimates. The statistics obtained under KNODE-MHE and KNODE-UKF and comparisons with MHE and UKF are depicted on the right of Fig. \ref{fig:crazyflie_results}. By computing the median of the overall MSE, it is observed that KNODE-UKF and KNODE-MHE outperforms their non-learning counterparts by 26.6\% and 11.9\% respectively. It is also observed that MHE generally outperforms UKF, and this trend is similar to that from the results of the other state estimation tasks. In general, the scalability of the proposed framework depends on the type of residual dynamics present within the system, as this influences the complexity of the NODE model. In \cite{chee2022knode}, it has been shown that a relatively small model is required to model the residual dynamics for a high-dimensional quadcopter.
\vspace{-0.1cm}
\section{CONCLUSION AND FUTURE WORK}
\vspace{-0.1cm}
In this work, we present LEARNEST, a versatile framework in which we learn and integrate a knowledge-based, data-driven KNODE model into two state estimation algorithms. Simulation results and experiments with real-world data across various applications show that the proposed algorithms outperform their non-learning counterparts. This demonstrates the effectiveness of LEARNEST and validates the accuracy of the learning-enhanced state estimation algorithms. In future work, we plan to apply this framework on other model-based state estimation algorithms such as the particle filter.

\bibliographystyle{IEEEtran} 
\bibliography{IEEEabrv,refs} 

\addtolength{\textheight}{-12cm}   





\end{document}